# Environmental Slow AI: Design Principles for Generative Systems

**Vanessa Utz**[1]

## Abstract

Generative AI (genAI) systems produce cultural artefacts at scale, but they also reflect embedded cultural values through their design. Once identified, these values become open to deliberate reshaping. This position paper examines the maximalist values of current generative AI through an environmental humanities tradition and proposes design principles in which environmental sustainability serves as the core value instead. The principles are developed under the umbrella of *Slow AI*, a term that already circulates across several distinct research and practice programs. Five design principles are articulated (restraint, sufficiency, selectivity over retention, material visibility, and friction as affordance), each of them illustrated against the current design of widely deployed systems. Each principle operates at two levels: a design implementation, and an interpretive layer at which users and developers are prompted toward reflective engagement with the system. Together these principles extend human agency by restoring decisions that frictionless defaults have silently removed and do so by building interpretive reflection into design.

## 1. Introduction

Current generative AI expresses a consistent commitment to maximalism (Evans, 2024) and zero-friction (Xu et al., 2026). Midjourney returns four images per prompt by default (Midjourney, n.d.) and its Discord integration provides increased accessibility. The newly released ChatGPT Image 2 returns up to eight images per prompt in thinking mode (Singh & Okcular, 2026), doubling Midjourney's default while adding reasoning steps that increase compute per invocation. Further, ChatGPT presents a perpetually available prompt field and a single-click regenerate button. These design choices lead to increased usage amongst the user base. Survey data on text-to-image users has shown individual outputs of 2,000 to 5,000 images per week among power users when these systems first became widely available (Utz & DiPaola, 2023).

GenAI's relationship with our culture therefore exists in two ways: 1) genAI systems are producers of cultural artefacts: images, text, audio, and video generated at scale and deployed across creative, professional, and everyday contexts, and 2) the design of these systems also embed cultural values in their material and interface choices: the default output form, any limitations on system use, data storage defaults, as well as the cost of use. These design choices are expressions of a view about how genAI should fit into our world, and they are available to analysis in the same way as the output they produce.

This paper takes the position that the maximalist framing of genAI design is an expression of a cultural value rather than a technical necessity. Due to this status quo, the environmental consequences of genAI are substantial and continue to grow (Berthelot et al., 2025; Falk et al., 2025). Inference-phase energy demand has reached or overtaken training emissions in cumulative terms for widely deployed models (Luccioni & Hernandez-Garcia, 2023; Chien et al., 2023; Bashir et al., 2024). Operational and supply-chain water demand has now been systematically quantified across training and inference (Li et al., 2025). Data centre associated water consumption is putting stress on communities. For instance, 6 percent of West Des Moines' city water was consumed by GPT-4 training in a single month (Hosseini et al., 2025). Storage and retention costs associated with default-saved generative outputs remain under-studied but scale with the user base (Utz & DiPaola, 2023). Hardware manufacturing accounts for the majority of toxicity and resource-depletion impacts in lifecycle assessments of generative AI, with use-phase activity contributing the remainder of climate impact (Falk et al., 2025). The lifecycle picture is therefore not reducible to inference-phase energy alone, even where inference dominates ongoing consumption. These consequences are not implementation bugs that await optimisation. They are effects of the value system in which the technology has been designed. This work therefore argues that environmental protection and sustainability can serve as an alternative cultural commitment around which genAI design can be developed. The design principles of this new commitment are presented here as a contribution to Slow AI.

---

[1] Simon Fraser University, Surrey, BC, Canada. Correspondence to: Vanessa Utz, vutz@sfu.ca.

Slow AI, as developed here, names the design-research program organised around environmental protection and sustainability as its primary commitments. It is important to note that the program is positive rather than subtractive. Instead of proposing less AI, it proposes that invocation be treated as a meaningful design decision rather than pre-answered by frictionless defaults.

## 2. Slow AI and Adjacent Programs

The term *Slow AI* extends Hallnäs and Redström's (2001) *Slow Technology*, which proposed reflection rather than efficiency as the central purpose of computing design and treated slowness as a design surface through which users come to understand a technology and its place in their lives (see Odom et al., 2012, for the field's later development). A parallel lineage in industrial design (Strauss & Fuad-Luke, 2008) developed slow design as a principles-based reflective tool for sustainability. Crampton (2020) developed the AI specification of this work, organising Slow AI around three imperatives (Think, Resist, Act Local) and framing AI as a technology that demands mindful adoption rather than automatic deployment.

The term has since circulated across several research and practice communities. Huggett (2024) connected it to archaeological practice through the Hallnäs and Redström lineage. Illingworth (2025) developed a pedagogical Slow AI concerned with assessment integrity and reflection in higher education. Further applications exist in creativity research and decolonial design practice (Glaveanu & Blackwell, 2025; Beghetto, 2023; Fernández Mora et al., 2024). These threads share a single design approach: restoring deliberation as a design value against orthodoxies that treat friction as cost. They differ on which cultural value organises that move, anchoring it variously in governance and place-based co-production, archaeological practice, pedagogical integrity, creativity, or decolonial design.

Parallel conversations around Sustainable AI have taken place across CS research. For example, Green AI (Schwartz et al., 2020) and Frugal AI (Arga et al., 2025) are sister initiatives to our proposal sharing the premise that current development trajectories impose unsustainable resource demands. Green AI advocates elevating efficiency to a co-equal evaluation criterion alongside accuracy in research practice. Frugal AI extends this orientation to system design, drawing on model compression, hardware optimisation, and eco-design, with attention to the economic and regulatory conditions of development. Both programs target supply-side optimisation: Green AI addresses research practice, while Frugal AI addresses the engineering of systems. Neither of them deliberately focuses on the use phase of genAI systems and user engagement. However, the programs are complementary to our environmental Slow AI. Supply-side optimisation does not constrain expanded use, and use-phase commitments do not produce more efficient models. The environmental case for genAI design therefore requires both, a point returned to under rebound (Section 6).

Our environmental Slow AI proposal sits at the intersection of these research fields. It draws on Slow Technology for its design vocabulary of restraint, sufficiency, and reflection, and on the broader Sustainable AI conversation for its diagnostic premise. The contribution of this work is an environmental Slow AI operating at the user interface, with environmental concern treated as the cultural value organising the design.

## 3. Slow Violence and Intergenerational Equity

The term *slow violence* refers to gradual, deferred, and dispersed harm that escapes the aesthetic and political forms through which violence is ordinarily recognised (Nixon, 2011; Pain & Cahill, 2022). Climate change, soil degradation, and the health burdens of toxic dumping all exhibit this structure: harm accumulates over decades or generations, occurs in places separated from its causes, and resists the framing demanded by news cycles and political debate. GenAI at scale exhibits this structure at the inference layer. The user sees a chat interface. The data centre, cooling infrastructure, and the semiconductor fabrication that produced the required chips remain invisible. The harm is spatially, temporally, and representationally dispersed.

Nixon's underlying argument is not about speed but about perception. Slow violence is violence that escapes the narrative forms by which harm is ordinarily made visible. The political task Nixon identifies is making slow harm legible (this task has been taken up across humanities work, e.g., Perkins, 2022). The *Material Visibility* principle developed in Section 4 is the design-layer answer to that task. Visibility of inference cost at the interface is a direct operationalisation of Nixon's framework at the appropriate scale.

Slow violence works through decoupling: the cause (a prompt) is separated from consequence (resource consumption and waste production distributed across localities and generations). Nixon argues this decoupling is structurally produced and politically useful, in that it insulates harm generators from accountability. The *Restraint and Friction principles* developed in Section 4 function as recoupling interventions. Each invocation of a generative model becomes a site at which deferred and dispersed consequences re-enter the moment of decision. The temporal structure that allows slow violence to evade accountability is deployed in reverse to reassert it.

Nixon argues that temporality itself is politically constructed. *Fast time* (the time of product launches, benchmark rankings, and quarterly reports) is culturally legible. *Slow time* (the time of accumulation, inheritance, and deferred cost) on the other hand is structurally suppressed. Slow AI, in the environmental specification developed here, does not propose that generative systems should be used more slowly. It proposes that slow time is a mode of attention to phenomena that fast-time framings have obscured.

Nixon's framework is mainly used analytically here and serves to identify patterns and underlying structures in the harms caused by the current design approach to genAI. On the other hand, Brown Weiss's framework of intergenerational equity (defined as the principle that the planet is held in trust across generations, with each generation owing the next a comparable inheritance) provides the ethical motivation to act. Future generations hold equal standing as beneficiaries of the planetary commons (defined as the atmosphere, biosphere, freshwater systems, and the broader resource base held collectively across generations rather than assignable to any one party), and practices that externalise cost to them violate a duty of conservation that present generations hold in trust (Brown Weiss, 1989; Brown Weiss, 1990). Brown Weiss specifies this duty through three principles: 1) conservation of options (preserving diversity of resources and choice), 2) conservation of quality (passing on the planet in no worse condition than received), and 3) conservation of access (equitable access to its benefits). Through this lens, the current maximalist paradigm is intergenerationally unjust, and the Section 4 design principles operationalise these duties in terms of genAI: *Sufficiency* and *Selectivity* preserve options and quality at the level of computational and storage commons. An environmental reframing of Slow AI treats intergenerational cost accounting as a fundamental design input.

## 4. Design Principles

We present five design principles to reshape genAI. Each principle operates at two levels. The first is the material layer: a specific design commitment that changes what the system does or defaults to. The second is the interpretive layer: a moment at which the design choice demands reflective engagement from a user or a developer. The interpretive layer is a way the humanistic analysis of genAI (Section 3) connects to the design principles. Each principle is developed below through contrast with current system behaviour.

*Principle 1 - Restraint.* The choice to invoke a generative model is itself a design decision. Current consumer interfaces render invocation frictionless. ChatGPT presents an always-available prompt field, Gemini integrates into Google Workspace so that users encounter generative suggestions whether or not they sought them, and Anthropic's Claude has rolled out comparable integrations into Chrome and the Microsoft Office suite, embedding generative invocation directly into browsing and document workflows (Anthropic, 2025). The question "should I use this model for this task?" is pre-answered by the design of the interface. Empirical work further supports this. Qiao et al. (2025) found that users made systematically poor decisions about whether to engage generative AI assistance, characterising two gaps: a gulf of impatience, where users refuse assistance they would benefit from, and a gulf of overreliance, where users accept assistance that produces worse outcomes than working unaided. *Restraint* treats non-use as a first-class option. This does not mean building systems that discourage their own use. It means systems that present the choice of whether to engage them as a meaningful one, with contextual support for recognising when a conventional tool, or no tool at all, serves the task better. The agency restored is the choice itself. *Restraint* demands that the user reads the task at hand and judge whether it warrants algorithmic response, and it requires the developer to have made a prior choice about which decision points count as meaningful and worth surfacing.

*Principle 2 - Sufficiency.* Outputs should be sized to task, not to demonstration. Abundance-paradigm systems reward maximal output. ChatGPT Image 2 defaults to 1024x1024 (Singh & Okcular, 2026), ChatGPT returns multi-paragraph responses to factual questions answerable in a sentence and Midjourney generates four variants per prompt because the system is designed around exploration rather than selection. This current work treats fitness-for-purpose as the quality target. Inference cost scales with output size, so fitted outputs are lower-cost outputs. The cultural implication is that the system models a better relation to the user's actual need. *Sufficiency* operates materially as a constraint on default output size and computational expenditure, and interpretively as a paired judgement: the developer specifies what fitness between output and task means at design time, and the user reads the resulting output for whether it meets the task or exceeds it. This resource-economy rationale parallels Frugal AI's minimal-footprint commitment (Arga et al., 2025), but our *Sufficiency* principle places the constraint at the user interface and ties it to interpretive practice rather than to model-side engineering.

*Principle 3 - Selectivity over Retention.* Current systems default to retention. For instance, Midjourney's public feed preserves every generation indefinitely (Midjourney, n.d.); ChatGPT retains conversation history as the default user setting, with deletion requiring affirmative user action (OpenAI, n.d.). The retention default is partly responsive to user request, but it is also a reflection of values: storage is treated as cheap, its cost is externalised to data-centre operations, and nothing in the interface makes curation the natural next action. A Slow AI specification inverts this default. Outputs not affirmatively retained by the user within a defined window (for instance, a session boundary or a fixed expiry of days) are purged from system storage. Materially, the principle commits the system to opt-in retention; interpretively, it returns the curatorial act to the user as a deliberate choice rather than a passive accumulation. The consequence is reduced storage load and the user's relation to their own outputs becomes selective rather than accumulative. *Selectivity* returns to the user a curatorial act that retention had foreclosed, and it requires developers to treat defaults as statements about what the system considers worth preserving.

*Principle 4 - Material Visibility.* The material cost of inference should be visible at the point of interaction. Current systems expose little of this information to end users. HuggingFace supports optional carbon-emissions reporting on model cards (HuggingFace, n.d.) but widely deployed consumer genAI products do not currently expose per-invocation energy, water, or comparable cost figures at the point of use. Forms this could take include inline energy indicators, comparative figures against non-AI baselines, and water-use estimates contextualised to the user's locality. Quantifications are now emerging at the API level (e.g., EcoLogits, Rincé & Banse, 2025). The goal of these visible metrics is not to shame users into non-use but to make the unseen implications of decisions available to them, in the way nutrition labelling and carbon-footprint indicators on flights have made material registers available in other consumption domains. Materially, the principle commits the system to surfacing inference cost (energy, water, or comparable measures) at the interface where invocation occurs. Interpretively, *Material Visibility* is the principle at which Nixon's diagnostic apparatus enters design most directly: representing dispersed harm at an interface is itself an interpretive act, and the developer designing that representation is performing the work Nixon identifies as politically necessary. *Material Visibility* extends the price-tag rationale of Green AI (Schwartz et al., 2020) from research publication to user interface. While Schwartz et al. proposed making computational cost legible to researchers as an evaluation dimension, this principle makes it legible to users as a precondition for environmentally informed engagement.

*Principle 5 - Friction as Affordance.* Pauses, prompts, and confirmations can operate as affordances for reflection. This principle inherits directly from Hallnäs and Redström's (2001) work on Slow Technology. Current design paradigms treat friction as cost to be minimised. Slow AI, however, treats certain frictions as features rather than defects. This includes strategies such as structural pauses before high-cost operations and guidance tools that prompt the user to revise prompts rather than regenerate outputs. Materially, the principle commits the system to retaining specific frictions (pauses, confirmations, prompt-revisions) at points where the cost or consequence of an operation demands reflection. Where *Material Visibility* presents hidden costs to the user, *Friction* interrupts the action that produces it. The two principles operate at adjacent but distinct points in the interaction. Hallnäs and Redström argued that "slowness is a key factor that could bring forth, and make room for, reflection" (2001, p. 210).

These five principles do not exhaust the design space, and no single system needs to instantiate all of them. Their function is to specify what an environmentally grounded Slow AI commits to at both layers (design implementation and interpretive surface), such that subsequent empirical work has a clear target against which to build and evaluate.

## 5. Evaluation Implications

Evaluation practices in genAI reward the maximalism paradigm by default. Headline capability benchmarks such as MMLU (Hendrycks et al., 2021) and image generation metrics including FID (Heusel et al., 2017) and CLIPScore (Hessel et al., 2021) reward output count, visual fidelity, and prompt alignment. Model leaderboards that are organised around these metrics inherit and propagate these definitions of progress and success. Green AI proposed adding efficiency as a co-equal evaluation criterion (Schwartz et al., 2020), and several dedicated efficiency benchmarks have emerged since then. For example, HELM measures efficiency among seven core metrics (Liang et al., 2023) and the AI Energy Score leaderboard rates more than 166 models across 10 tasks (Luccioni, 2025). However, the cost figures remain researcher-facing rather than user-facing.

The principles in Section 4 imply additions to this existing evaluation ecosystem. For example, restraint rates (meaning the frequency at which a system models or recommends non-use where appropriate), and task-fitness measures (the relation between generation size and task requirement) could be developed as use-phase environmental metrics. Intergenerational cost accounting is methodologically harder to specify. Lifecycle assessment frameworks that are now being applied to AI systems (Falk et al., 2025; Arga et al., 2025) supply the methodological starting point, but extending lifecycle accounting to capture harm distributed across future generations remains an open question to be addressed in future work. Full development of these metrics is itself a research program and lies beyond the scope of this paper. Restraint rate is briefly sketched out below as an illustrative example, because it has no existing analogue in the benchmarking literature.

*Restraint Rate*: A curated benchmark of prompts classed by task type would form the basis. Prompts in the first class would have a conventional non-AI solution at comparable quality (e.g., a calculator for an arithmetic question). The second class would cover prompts where generative assistance offers marginal benefit, and the third would cover prompts where it offers clear benefit. Ground truth for the first class can be established by human annotators with relevant domain expertise. The second class would be harder to label, and annotator disagreement on these prompts should likely be retained in the benchmark, since the disagreement is itself informative about which cases are genuinely contested.

Restraint rate is then the proportion of first-class prompts for which the system either declines invocation, redirects the user to a conventional tool, or presents non-use as a recommended option. This is analogous to refusal-rate evaluation in safety benchmarking, where the field has accepted that not producing an output can be the correct system behaviour. Under Slow AI, the same evaluation infrastructure can be extended to environmental restraint as a parallel quality dimension.

These are additions rather than replacements to existing evaluation practices. A system achieving comparable capability at substantially lower material cost should count as better, not worse, under an evaluation framework that takes environmental harm seriously. The construction of these metrics is itself a research program that follows from the position this paper develops.

## 6. Limitations

The proposal of reframing a sustainable Slow AI, and designing for it, carries some limitations.

The first is the lifestyle critique inherited from the broader slow movement lineage. For instance, Slow Food has been criticised for what Thompson and Kumar (2021) describe as *consumer responsibilisation*: an ideological pattern in which environmental responsibility is shifted to individual consumers rather than addressed at the level of production systems. A Slow AI positioned primarily as an individual user virtue risks reproducing this pattern, making environmental responsibility the burden of users who can afford deliberation while leaving the infrastructure unchanged. However, the present proposal is that Slow AI operates at the layer of system design, not individual virtue. The design principles in Section 4 specify what system builders commit to, not what users must perform. The burden belongs where the design decisions are made.

The second is rebound. Even where material visibility and restraint encourage moderation at the individual interaction, aggregate demand may grow through the diffusion of generative AI into new populations and use contexts (Luccioni et al., 2025; Arga et al., 2025). Individual restraint does not automatically produce aggregate restraint. Slow AI as a design commitment requires pairing design-level commitments with infrastructural and governance commitments that fall outside the scope of this paper.

The third is greenwashing. Any design-research program organised around environmental commitment is vulnerable to co-optation: firms can adopt the vocabulary while preserving harmful practices. The specificity of the present proposal (named design principles, named evaluation metrics, and measurable material outcomes) is what distinguishes it from a branding exercise. Without operational rigour, the term could be absorbed into the paradigm it names to oppose.

## 7. Conclusion

Environmental sustainability is a coherent cultural value around which generative AI design can be organised, and Slow AI is an existing conversation within which that value can be specified. The contribution of this paper is the specification itself: an environmental Slow AI, operating at the inference and user interface layer, grounded in a slow violence diagnosis, an intergenerational equity normativity, and the slow technology lineage. The five design principles (restraint, sufficiency, selectivity over retention, material visibility, and friction as affordance) restore rather than reduce human agency, because each reintroduces a decision that frictionless defaults had silently removed. Each principle operates materially and interpretively, embedding a humanistic apparatus for reading environmental harm into design rather than applying it to design from the outside. We do not propose that users should use less AI, but that the use of generative AI should be treated as a meaningful act by design, accountable through evaluation, and specifiable in concrete terms at the user interface.

## Impact Statement

The environmental Slow AI proposal could have broad negative consequences if misapplied. For example, the proposal's vocabulary could be adopted without the substance, leading to potential greenwashing. The interpretive surface could highlight environmental cost to users without giving them agency to act on it, generating guilt rather than informed engagement. The restraint principle could be applied in ways that fall disproportionately on users without access to non-AI alternatives, while leaving infrastructural decisions untouched. The cultural-values framing risks being read as anti-progress and dismissed on those terms by audiences for whom environmental commitment registers as a constraint on innovation rather than a design value. None of these are reasons not to pursue the proposal, but they are reasons to pursue it with attention to who bears its costs and who designs its implementations.